\documentclass[journal,twoside]{IEEEtran}
\usepackage{amsfonts,bm,booktabs,cite,graphicx,hyperref,multirow,textcomp,url}
\usepackage{algorithm}
\usepackage{subfigure}
\usepackage{algpseudocode}
\usepackage{amsmath}
\usepackage{mathrsfs}
\usepackage{float}
\usepackage{color}
\usepackage{booktabs}
\usepackage{bbding}
\usepackage[flushleft]{threeparttable}

\makeatletter 
\renewcommand{\@thesubfigure}{\hskip\subfiglabelskip}
 \makeatother
\begin{document}
\title{DCVC-MV: Deep Contextual Multiview Video Compression with Efficient Inter-View Prediction}
\author{
	Xihua Sheng, \IEEEmembership{Member, IEEE},
	Yingwen Zhang,
	Long Xu, \IEEEmembership{Senior Member, IEEE}, \\
	Shiqi Wang, \IEEEmembership{Senior Member, IEEE}
\thanks{
Date of current version \today.\par 

X. Sheng, Y. Zhang, and S. Wang are with the Department of Computer Science, City University of Hong Kong, Hong Kong, China (e-mail: xihsheng@cityu.edu.hk;ywzhang26@um.cityu.edu.hk;shiqwang@cityu.edu.hk).\par
Long Xu is with the Faculty of Electrical Engineering and Computer Science, Ningbo University, Ningbo 315211, China (e-mail: lxu@nao.cas.cn). \par

Corresponding author: Shiqi Wang.\par
}
}

\markboth{IEEE Transactions on Multimedia}{DCVC-MV: Deep Contextual Multiview Video Compression with Efficient Inter-View Prediction}

\maketitle
\begin{abstract}
Multiview video is a key format for 3D applications such as free-viewpoint broadcasting and virtual reality, yet its large data volume poses significant challenges for efficient storage and transmission. As deep contextual video compression matures and moves toward standardization, extending such learned codecs to multiview scenarios has become essential for practical deployment---yet this direction remains largely unexplored.
In this paper, we propose DCVC-MV, a novel deep contextual multiview video compression framework that satisfies three fundamental requirements. First, it maintains backward compatibility, ensuring that the primary view's bitstream can be decoded independently by a single-view decoder without being affected by other views. Second, it supports random-access capability, enabling flexible switching between different viewing perspectives. Third, it effectively exploits inter-view correlations to achieve high compression efficiency. This is realized through four dedicated components: (1) an inter-view motion feature propagation method, which propagates decoded independent-view motion features as conditions to promote dependent-view motion encoding; (2) an inter-view motion conditional entropy model designed to learn motion conditional priors across views for more accurate probability estimation of motion latent representations; (3) an implicit inter-view context prediction method, which predicts inter-view contexts from low-resolution independent-view content features without explicit disparity estimation; and (4) an inter-view contextual conditional entropy model that learns contextual conditional priors across views to further enhance content compression.
Experimental results show that DCVC-MV delivers superior rate-distortion performance compared to the traditional MV-HEVC standard, the independent single-view video compression baseline, and the state-of-the-art implicit neural representation-based multiview video codec, while fully adhering to the structural and compatibility constraints of practical multiview systems. This work establishes a strong baseline that fulfills the critical requirements of real-world multiview applications, facilitating future research and standardization in this evolving field.

\end{abstract}
\IEEEpeerreviewmaketitle

\section{Introduction}
With the popularity of 3D immersive applications across diverse fields, multiview video has become a key media format for delivering realistic and interactive visual experiences. Captured by synchronized arrays of cameras surrounding a scene, it provides multiple overlapping viewpoints that enable functionalities such as free-viewpoint navigation and dynamic perspective switching. However, the large number of cameras leads to a significant surge in data volume, creating a pressing need for efficient compression to mitigate the associated storage and transmission costs.\par

Building upon mature single-view coding frameworks---originally designed to exploit spatial and temporal redundancies---multiview video coding (MVC) extends these architectures by incorporating inter-view prediction, thereby effectively capturing statistical dependencies across different viewpoints to achieve substantially higher compression efficiency. Correspondingly, international standards for MVC have been established as extensions of their single-view predecessors, such as the MVC extensions of H.264/AVC~\cite{vetro2011overview}, and H.265/HEVC~\cite{tech2015overview}. To meet the practical requirements of immersive applications, these standardized MVC architectures are generally designed around three key principles:
(1) \textbf{backward compatibility}, which ensures that the bitstream of the primary view can be decoded independently by a standard single-view decoder; (2) \textbf{random-access}, which allows flexible switching to any viewpoint or temporal point during playback; and (3) \textbf{efficient inter-view prediction}, which explicitly removes redundancies across views through techniques such as inter-view motion prediction~\cite{guo2006inter,sheng2025drfc,lee2010motion,gao2015encoder,schwarz20123d,schwarz2012inter,konieczny2010depth}, disparity compensation-based prediction~\cite{merkle2007efficient,shao2011asymmetric,kim2007fast,pan2015efficient,shen2010view},  and inter-view residual prediction~\cite{shen2010early}. These tools, operating in conjunction with conventional intra- and inter-prediction, form the core of modern standardized MVC solutions.\par

The rapid progress of deep learning-based video compression, particularly the deep contextual video compression (DCVC) series work~\cite{li2023neural,wang2024long,li2024neural,sheng2025bi,sheng2024vnvc,sheng2026fine}, has demonstrated performance comparable to or even surpassing that of conventional standards such as H.265/HEVC and H.266/VVC. Benefiting from its superior compression efficiency and end-to-end optimization capability, the DCVC architecture has attracted significant attention from standardization bodies and is currently under active development for practical deployment. Notable examples include the AVS-EEM~\cite{sheng2026recent} under development by the Audio Video coding Standard (AVS) group, and the MPAI-EEV~\cite{jia2023mpai} developed by the Moving Picture, Audio and Data Coding by Artificial Intelligence (MPAI) organization.\par
Extending the DCVC framework to multiview video is a critical and timely research direction, yet it introduces substantial technical challenges. The core difficulty lies in effectively integrating inter-view prediction into the deep contextual video coding framework while strictly preserving the three essential structural constraints of practical MVC systems: backward compatibility, random-access, and effective redundancy reduction across views. Successfully addressing these challenges would establish a deep contextual multiview video codec that is both high-performing and fully compatible with legacy system requirements, thereby bridging a key gap between neural video compression and real-world immersive applications, and offering direct contribution to ongoing standardization efforts.

\par
\begin{figure}[t]
  \centering
   \includegraphics[width=\linewidth]{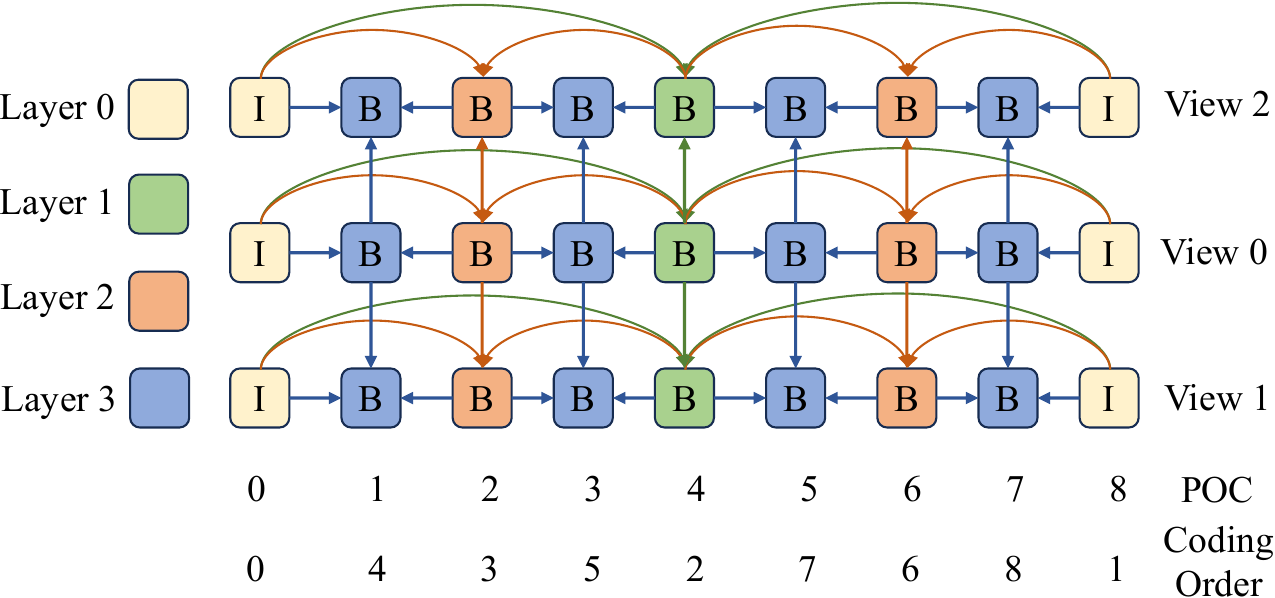}
      \caption{Proposed reference structure of the DCVC-MV framework. View0 serves as the independent view, with View1 and View2 being the dependent views.}
   \label{fig:GOP_structure}
\end{figure}
In this paper, we present DCVC-MV, a novel deep contextual multiview video compression framework designed to satisfy the three core requirements of practical multiview systems.
To achieve backward compatibility, the independent view is compressed using a bidirectional deep contextual single-view video codec, ensuring that its bitstream can be decoded independently.
As illustrated in Fig.~\ref{fig:GOP_structure}, to enable temporal and inter-view random-access, DCVC-MV employs a hierarchical bidirectional reference structure along the temporal axis, while in the view direction, the first frame of each group of pictures (GOP) in a dependent view is encoded independently---without referring to any prior temporal frames or other views---thereby supporting flexible viewpoint switching.
To realize efficient inter-view prediction, DCVC-MV introduces four dedicated coding components: an inter-view motion feature propagation  method (IVMFP), which propagates decoded bi-directional motion features as conditions to promote the motion encoding of the dependent-view video; an inter-view motion conditional entropy model (IVMCE), which learns motion conditional priors across views from bi-directional motion vectors and motion latent representations of independent-view videos to improve the probability estimation for the motion latent representations of dependent-view videos; an implicit inter-view context prediction method (IIVCP) , which predicts contexts across views from the decoded features with lower resolution of the independent-view videos without explicit disparity estimation; and an inter-view contextual conditional entropy model (IVCCE), which leverages the predicted inter-view contexts along with the content latent representations of independent-view videos to learn contextual conditional priors across views. \par

Our contributions are summarized as follows:
\begin{itemize}

    \item We propose DCVC-MV, a deep contextual multiview video coding framework that simultaneously ensures backward compatibility and random-access capability, while effectively reducing inter-view redundancy through learned cross-view prediction.

    \item We propose an inter-view motion feature propagation method and an inter-view motion conditional entropy model, which can effectively leverage  motion correlation across views. 

	\item We propose an implicit inter-view context prediction method and an inter-view contextual conditional entropy model, which can better leverage content correlation across views.
    
\end{itemize}
Experimental results show that DCVC-MV achieves superior rate-distortion performance compared to the conventional MV-HEVC standard, the independent single-view video compression baseline, and the state-of-the-art implicit neural representation-based multiview video codec.

\section{Related Work}\label{sec:related_work}

\subsection{Traditional Multiview Video Coding}
Research into traditional multiview video coding spans nearly four decades. A foundational contribution was made by Lukacs et al.~\cite{lukacs1986predictive} in 1986, introducing a multiview image coding technique that employed disparity-compensated prediction to leverage correlation across different views. 
Standardization efforts in multiview coding were initiated by the 1996 update to H.262/MPEG‑2~\cite{okubo1995mpeg,aramvith2000mpeg},  where the left view is regarded as a base-view and forms a self‑contained, standards‑compliant bitstream, and the right view is encoded as an enhancement layer that predicts from the base view.
This architecture set the stage for multiview extension of the H.264/AVC standard~\cite{okubo1995mpeg,aramvith2000mpeg}. This extension introduced a hierarchical B‑frame structure spanning all views, along with advanced block‑based motion and disparity estimation, to jointly exploit temporal and inter‑view correlations.
Further improvements were introduced to enhance efficiency and reduce complexity within this standardized framework. Shen et al.~\cite{shen2010view} proposed to reduce the computational load of motion and disparity estimation methods through mode-complexity analysis and adaptive prediction. Guo et al.~\cite{guo2006inter} contributed a direct mode for inter-view prediction, allowing motion vectors of macroblocks to be inferred from already‑coded view data. 
A major leap occurred in 2012 is the development of  3D‑HEVC and MV‑HEVC standards~\cite{muller20133d,tech2015overview}. Among these standards,  lots of works~\cite{schwarz2012inter,chen2014motion,kang2014low,gao2025two,zhang2014low} focused on developing advanced inter-view prediction methods to improve coding efficiency. For example, Kang et al.~\cite{kang2014low} proposed to derive predicted disparity information from motion information of temporally and spatially neighboring blocks using a neighboring-block disparity vector prediction method, which eliminates needs for depth-layer dependencies.  Zhang et al.~\cite{zhang2014low} proposed to improve residual prediction accuracy through motion alignment across views.
\begin{figure*}[t]
  \centering
   \includegraphics[width=0.95\linewidth]{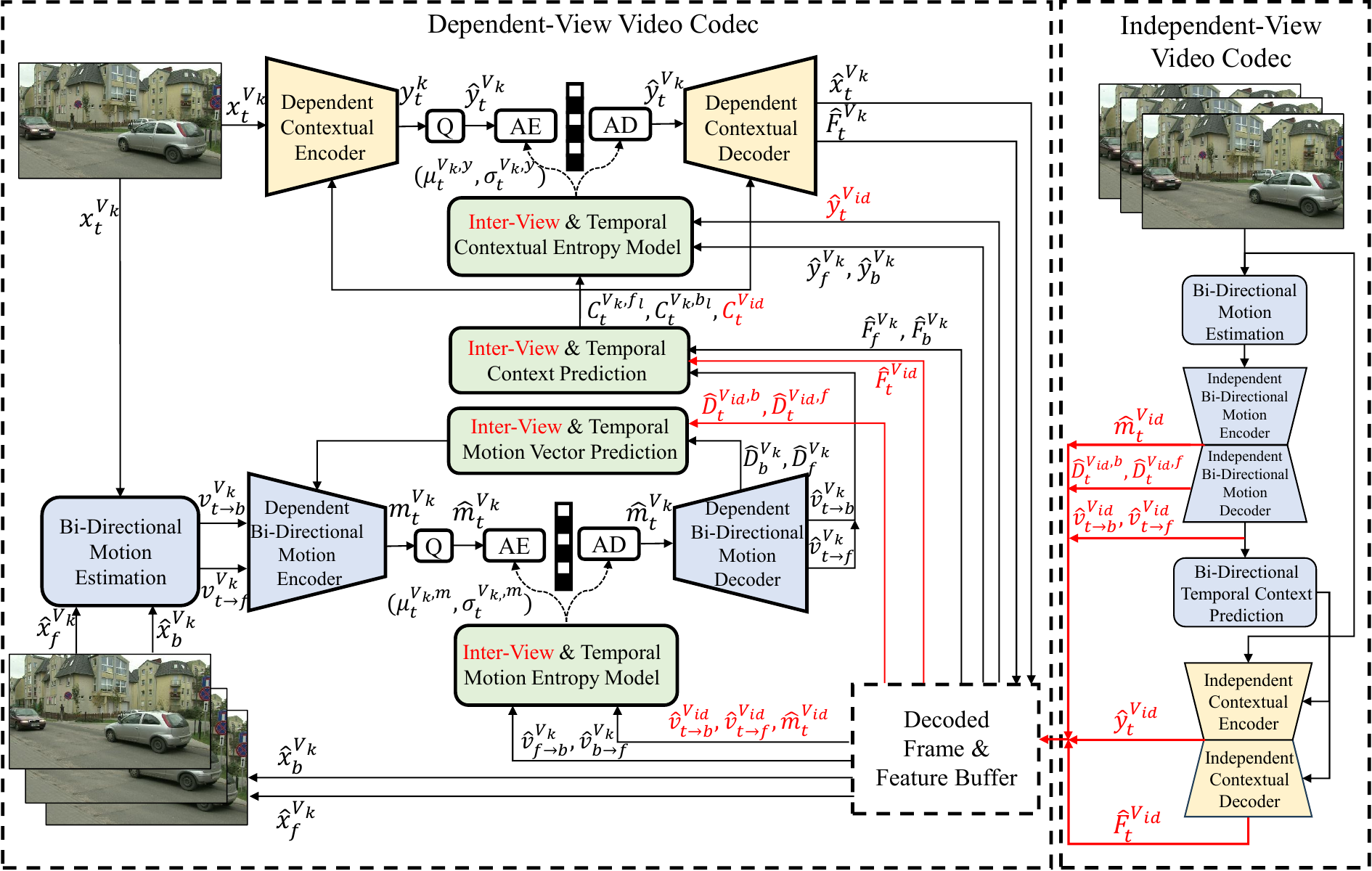}
      \caption{Overview of our deep contextual multiview video compression framework. 
}
   \label{fig:framework}
\end{figure*}

\subsection{End-to-End Deep Multiview Image and Video Compression}
For end-to-end deep multiview image compression, Liu et al.~\cite{liu2019dsic} introduced an autoencoder‑based stereo compression network that enables learnable feature-domain disparity-compensated prediction for the second view, augmented with an inter-view entropy model to reduce latent redundancy across views. Deng et al.~\cite{deng2021deep} proposed a homography‑based alignment scheme, compressing the prediction residual with a dual entropy model. Liu et al.~\cite{liu2024bidirectional}  proposed to use a 3D convolution to capture local features of stereo images and incorporate bi-directional attention blocks to exploit global features. More recently, Yang et al.~\cite{yang2024ficnet} proposed a learned multiview image codec equipped with a differentiable depth‑estimation method and a multi‑level fusion module to boost reconstruction fidelity. \par

For end-to-end deep multiview video compression, Chen et al.~\cite{chen2022lsvc} proposed a learned stereo video compression method for autonomous driving scenes. This method uses disparity compensation for inter‑view reference, combining motion compensation for temporal prediction for residual compression. Hou et al.~\cite{hou2024low} later designed a stereo video codec with a joint cross‑view prior model and a bi-directional feature shifting method for parallel encoding. Although these methods deliver competitive compression performance, their progressive or parallel architectures require re‑encoding the independent view and lack backward compatibility, making them ill‑suited for standards‑conformant multiview streaming. More recently, implicit neural representation (INR) has been introduced for multiview video coding~\cite{zhang2024efficient,zheng2024hpc, zhu2025implicit,shin2025neural,kwan2024immersive,gao2025bvi} is used for multiview video coding. For example, Zhu et al.~\cite{zhu2025implicit} proposed to map view and time indices into an INR to generate implicit representations. INR-based methods are advantageous for fast decoding. However, they typically lack support for random-access, necessitate extensive per-scene optimization during encoding, and are not compatible with conventional structured prediction frameworks. These limitations hinder their integration into emerging end-to-end learned video coding standards. In contrast, our work aims to develop a multiview video codec that simultaneously satisfies backward compatibility, random-access, and efficient inter-view prediction, thereby offering a practical, standards-ready solution. By adhering to these core requirements, our framework can directly contribute to ongoing standardization efforts such as AVS‑EEM~\cite{sheng2026recent} and MPAI‑EEV~\cite{jia2023mpai}, bridging the gap between learned video compression and real‑world immersive video applications.\par

\section{Proposed Framework}\label{sec:framework}
Figure~\ref{fig:framework} presents an overview of our proposed DCVC-MV framework. The framework consists of two core components: an independent-view video codec for compressing the primary view frames ($x_0^{V_{id}}, x_1^{V_{id}}, \cdots, x_N^{V_{id}}$), and a dependent-view video codec for compressing the frames of other views ($x_t^{V_k}, x_1^{V_k}, \cdots, x_N^{V_k}$, where $k$ denotes the view index).

\subsection{Independent-View Video Codec}
To achieve backward compatibility, the independent-view video codec in DCVC-MV encodes the primary view sequence as a self-contained bitstream, fully decodable by a single-view decoder. Simultaneously, to support temporal random-access, the codec employs a hierarchical bi-directional reference structure along the temporal axis, as illustrated in Fig.~\ref{fig:GOP_structure}. This codec integrates a suite of advanced compression methods, including bi-directional motion estimation, motion difference compression, motion entropy modeling, temporal context mining, contextual compression, and temporal contextual entropy modeling. Detailed implementations of these components are described in~\cite{sheng2025bi}. During compression, this codec generates several decoded representations that are subsequently leveraged to assist dependent-view encoding, including: decoded content feature $\hat{F}_t^{V_{id}}$, content latent representation $\hat{y}_t^{V_{id}}$, motion features ($\hat{D}_t^{V_{id},f}$, $\hat{D}_t^{V_{id},b}$), motion vectors ($\hat{v}_{t\rightarrow f}^{V_{id}}$, $\hat{v}_{t\rightarrow b}^{V_{id}}$), and motion latent representation $\hat{m}_t^{V_{id}}$.
\begin{figure}[t]
  \centering
   \includegraphics[width=\linewidth]{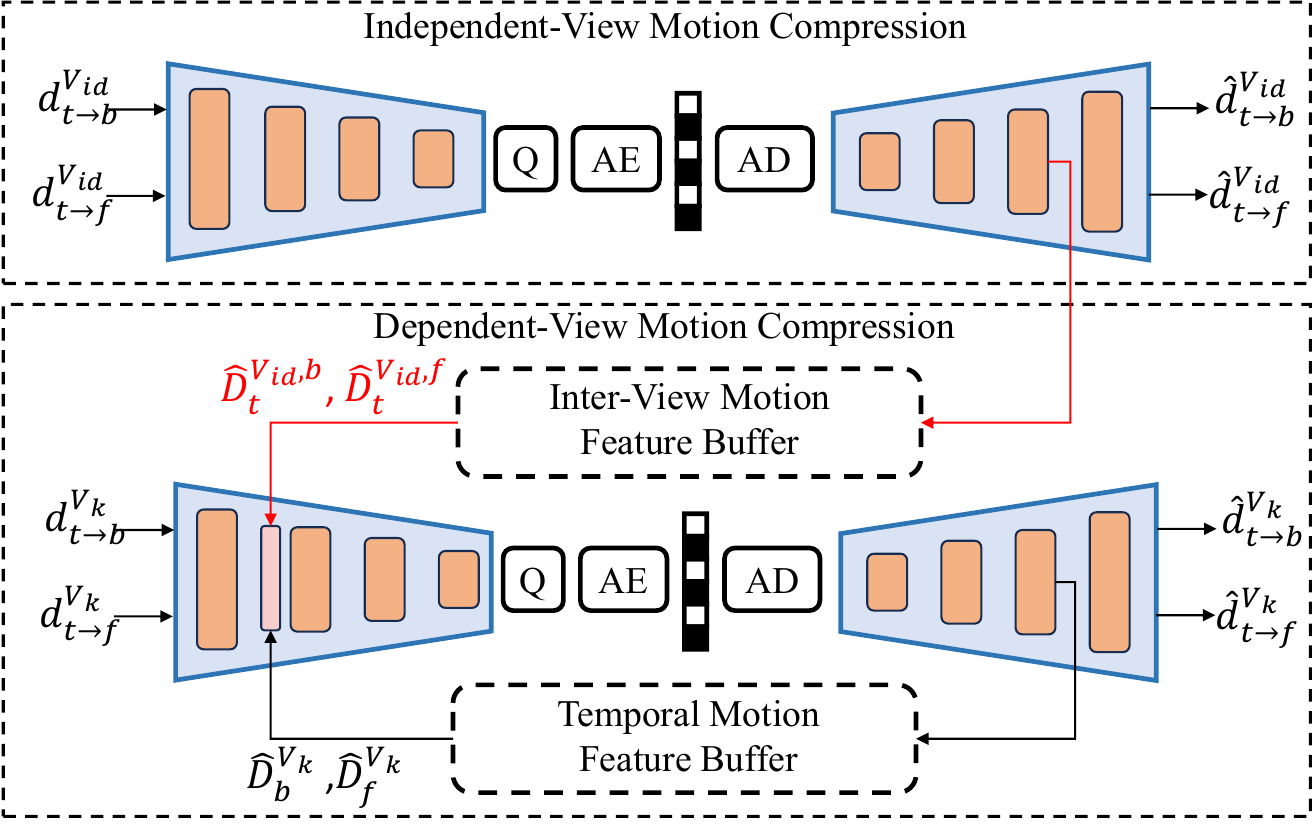}
      \caption{Motion vector compression with proposed inter-view motion feature propagation method. 
}
   \label{fig:IVMVP}
\end{figure}

\subsection{Dependent-View Video Codec}
Conditioned on the decoded representations of independent-view videos, a dependent-view video codec is designed for the dependent-view videos. To ensure random-access across different views, the first frame of each group of pictures (GOP) within a dependent view is encoded independently—without referring to any prior temporal frames or other views. For the remaining frames, a hierarchical bidirectional reference structure is adopted along the temporal axis, as illustrated in Fig.~\ref{fig:GOP_structure}, thereby achieving both inter-view and temporal random-access. To further enable efficient inter-view prediction and reduce cross-view redundancy, the dependent-view codec builds upon the temporal prediction framework of the independent-view codec and introduces four dedicated inter-view coding components: an inter-view motion feature propagation method, an inter-view motion conditional entropy model, an implicit inter-view context prediction method, and an inter-view contextual conditional entropy model. These components collectively leverage the decoded representations from the independent view to enhance the compression efficiency of dependent views.
\begin{figure}[t]
  \centering
   \includegraphics[width=0.8\linewidth]{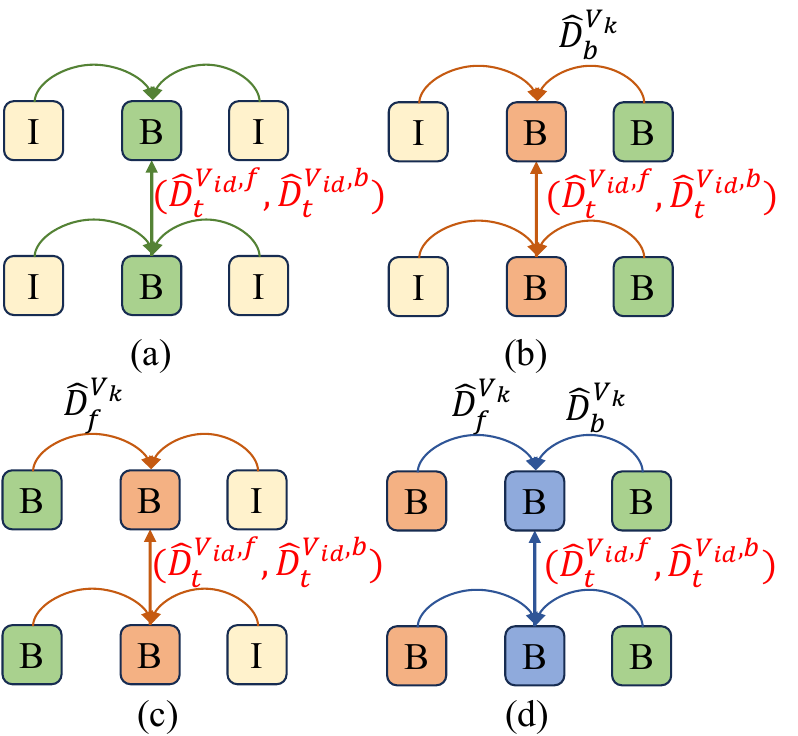}
      \caption{Inter-view motion feature fusion for different types of reference frames. 
}
   \label{fig:reference_propagation}
\end{figure}

\subsubsection{Inter-View Motion Feature Propagation}\label{IVMFP}
Compressing bi-directional motion vectors needs high bit rate costs in random-access configurations. To address this, our dependent view video codec first derives temporal motion prediction. Specifically, we perform bi-direction motion estimation to estimate bi-directional motion vectors ($\hat{v}_{b\rightarrow f}^{V_k}$,$\hat{v}_{f\rightarrow b}^{V_k}$) between bi-directional reference frames ($\hat{x}_f^{V_k}$, $\hat{x}_b^{V_k}$). We compute the motion vector differences ($d_{t\rightarrow f}^{V_k}$, $d_{t\rightarrow b}^{V_k}$) are then computed between bi-directional motion vectors ($v_{t\rightarrow f}^{V_k}$, $v_{t\rightarrow b}^{V_k}$) and their scaled motion predictions ($\frac{v_{b\rightarrow f}^{V_k}}{2}$, $\frac{v_{f\rightarrow b}^{V_k}}{2}$). These differences are encoded and decoded via a motion difference auto-encoder as illustrated in Fig.~\ref{fig:IVMVP}. Furthermore, we employ a feature-based motion difference information propagation method~\cite{sheng2025bi}, which regards the decoded temporal motion difference features $(\hat{d}_f^{V_k}, \hat{d}_b^{V_k})$ as conditions to improve the encoding of the current frame's motion difference. \par

To further exploit inter-view motion similarity, we propose an inter-view motion feature propagation method. The decoded bi-directional motion difference features $(\hat{D}_t^{V_{id},f}, \hat{D}_t^{V_{id},b})$ from the corresponding independent-view frame $x_t^{V_{id}}$ (of resolution $H/2 \times W/2$) are stored in the decoded motion feature buffer. When compressing the dependent-view frames $x_t^{V_k}$, these inter-view motion features are provided as additional conditioning to the motion encoder, enabling it to reduce cross-view motion redundancy.\par

We propose to fuse the inter-view and temporal motion features adaptively using dedicated motion feature fusion adaptors ($f_{\theta}$), implemented with depth-wise blocks~\cite{li2023neural}. The fusion strategy is determined by the frame types of temporal references. If the bi-directional references are I-frames, as illustrated in Fig.~\ref{fig:reference_propagation}(a), we only fuse the motion difference features across views ($\hat{D}_t^{V_{id}, f}$, $\hat{D}_t^{V_{id}, b}$) using $f_{{\theta}_0}$:
\begin{equation}
    D_{out}= f_{\theta_0}(D_{in}||\hat{D}_t^{V_{id}, f}||\hat{D}_t^{V_{id}, b}),
\end{equation}
where  ``$||$" denotes channel-wise concatenation, while $D_{out}$ and $D_{in}$ represent the output and input  features of the fusion adaptor, respectively.
If the backward reference is B-frame and the forward reference is I-frame, as shown in Fig.~\ref{fig:reference_propagation}(b),  we fuse the motion difference features across views ($\hat{D}_t^{V_{id}, f}$, $\hat{D}_t^{V_{id}, b}$) and the backward motion difference feature $D_b^{V_k}$ using $f_{\theta_1}$:
\begin{equation}
    D_{out} = f_{\theta_1}(D_{in}||\hat{D}_t^{V_{id}, f}||\hat{D}_t^{V_{id}, b}||\hat{D}_b^{V_k}).
\end{equation}
If the backward reference is I-frame and the forward reference is B-frame, as shown in Fig.~\ref{fig:reference_propagation}(c), we fuse the motion difference features across views ($\hat{D}_t^{V_{id}, f}$, $\hat{D}_t^{V_{id}, b}$) and the forward temporal motion difference feature $D_f^{V_k}$ using $f_{\theta_2}$:
\begin{equation}
    D_{out} = f_{\theta_2}(D_{in}||\hat{D}_t^{V_{id}, f}||\hat{D}_t^{V_{id}, b}||\hat{D}_f^{V_k}).
\end{equation}
If the bi-directional references are both B-frames, as shown in Fig.~\ref{fig:reference_propagation}(d),  we fuse the motion difference features across views ($\hat{D}_t^{V_{id}, f}$, $\hat{D}_t^{V_{id}, b}$) and the bi-directional motion difference features ($\hat{D}_f^{V_k}$, $\hat{D}_b^{V_k}$) using $f_{\theta_3}$:
\begin{equation}
    D_{out} = f_{\theta_3}(D_{in}||\hat{D}_t^{V_{id}, f}||\hat{D}_t^{V_{id}, b}||\hat{D}_f^{V_k}||\hat{D}_b^{V_k}).
\end{equation}
\begin{figure}[t]
  \centering
   \includegraphics[width=\linewidth]{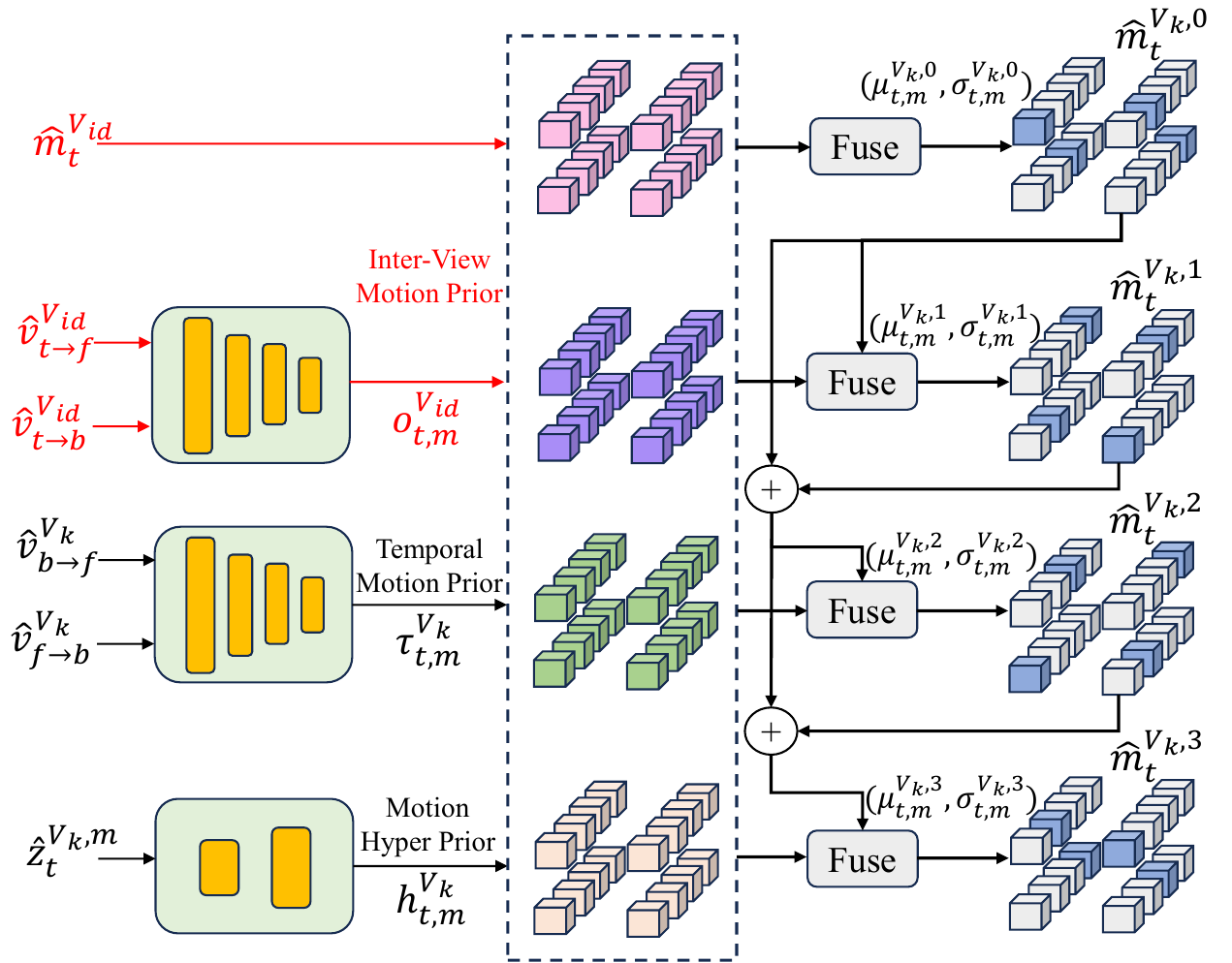}
      \caption{Motion conditional entropy model with proposed inter-view motion conditional priors. 
}
   \label{fig:MotionPrior}
\end{figure}

\subsubsection{Inter-View Motion Conditional Entropy Model}\label{IVMCM}
To more effectively exploit motion correlation across different views, we propose an inter-view motion conditional entropy model for dependent-view videos with inter-view motion conditional priors. We design an inter‑view motion prior extractor $f_{\psi}$, as illustrated in Fig.~\ref{fig:MotionPrior}, which consists of non‑linear activation functions and convolutional layers to predict an inter‑view motion conditional prior $o_{t,m}^{V_{id}}$ from the independent-view motion vectors $\hat{v}_{t\rightarrow f}^{V{id}}$ and $\hat{v}_{t\rightarrow b}^{V{id}}$:
\begin{equation}
o_{t,m}^{V_{id}} = f_{\psi}\bigl(\hat{v}_{t\rightarrow f}^{V_{id}} || \hat{v}_{t\rightarrow b}^{V_{id}}\bigr),
\end{equation}
where ``$||$" denotes channel-wise concatenation. The extracted inter-view motion prior $o_t^{V_{id}}$ shares the same spatial dimensions as the motion latent representation $\hat{m}_t^{V_k}$.
Furthermore, we also propagate the motion latent representation $\hat{m}_t^{V_{id}}$ of the independent view as another motion conditional prior across views.

The inter‑view motion conditional priors are integrated with the motion hyper prior $h_{t,m}^{V_k}$, and the temporal motion prior $\tau_{t,m}^{V_k}$. This combined prior information is subsequently fed into a spatial context model with quadtree partition~\cite{li2023neural} to estimate the distribution parameters $\mu_{t,m}^{V_k,s}$ and $\sigma_{t,m}^{V_k,s}$ of the $s^{\text{th}}$ dependent‑view motion latent segment $\hat{m}_t^{V_k,s}$:
\begin{equation}
p\left(\hat{m}_t^{V_k,s} \mid  o_{t,m}^{V_{id}}, \hat{m}_t^{V_{id}}, \tau_{t,m}^{V_k}, h_{t,m}^{V_k}, \hat{m}_t^{V_k, <s} \right) \sim \mathcal{L}\left(\mu_{t,m}^{V_k,s}, \sigma_{t,m}^{V_k,s}\right),
\end{equation}
where $\mathcal{L}\left(\mu, \sigma\right)$ denotes a Laplacian distribution and $\hat{m}_t^{V_k, <s}$ denotes previously compressed motion latent segments. 
\begin{figure}[t]
  \centering
   \includegraphics[width=\linewidth]{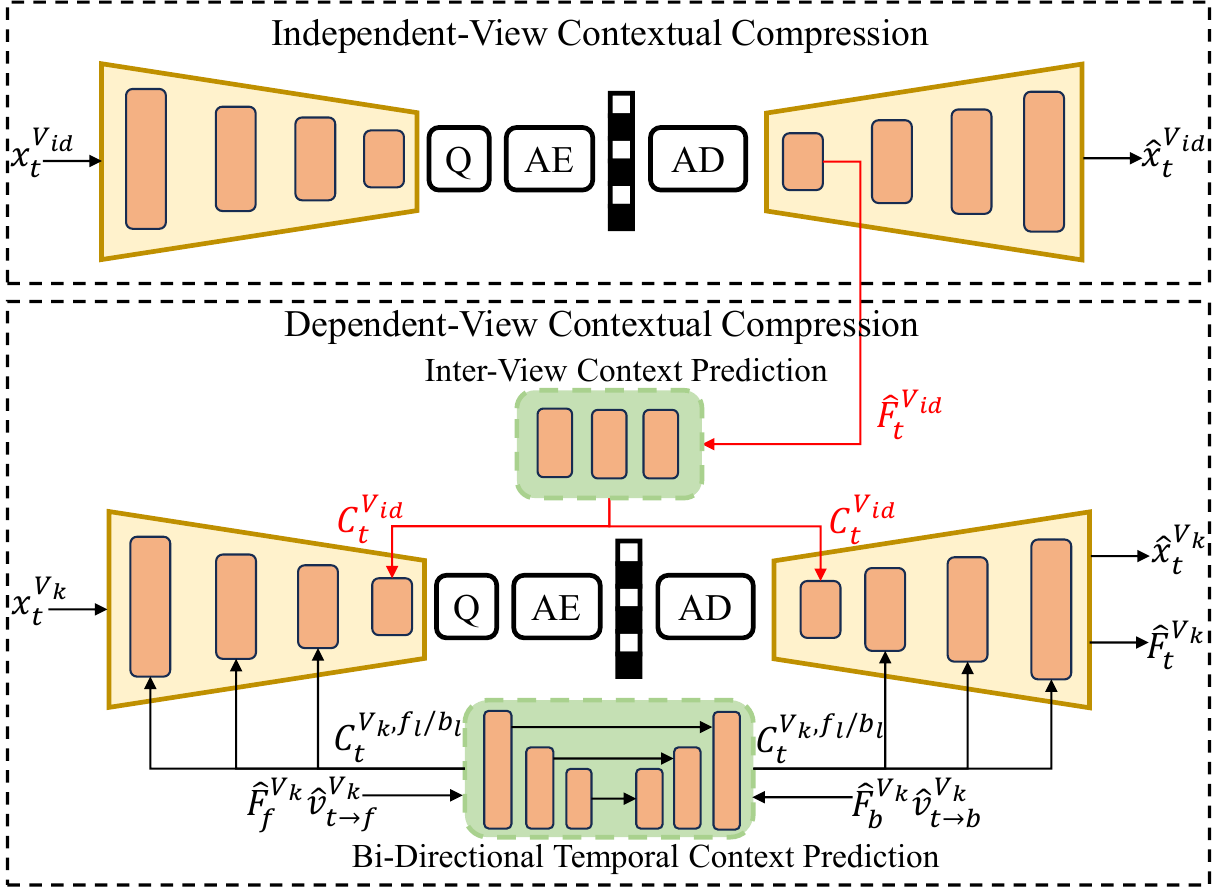}
      \caption{Contextual compression with proposed implicit inter-view context prediction method.
}
   \label{fig:context_prediction}
\end{figure}
\subsubsection{Implicit Inter-View Context Prediction}\label{IIVCP}
Significant content correlations between independent and dependent views present substantial opportunities for coding efficiency improvements.  Traditional techniques  estimate disparity fields explicitly and perform disparity compensation~\cite{shen2010view,pan2015efficient,kim2007fast,lei2022disparity,flierl2007motion,zhai2022disparity}.  However, in random-access configurations, bi‑directional learned video codecs already incur considerable computational overhead for motion estimation. Incorporating explicit disparity calculation would further escalate complexity, hindering practical implementation.
Therefore, we introduce an implicit inter‑view context prediction method without explicit disparity estimation.  Drawing on observations that downsampling can partially mitigate alignment errors~\cite{sheng2022temporal,sheng2024spatial}, as depicted in Fig.~\ref{fig:context_prediction}, our method propagates inter-view content information using a downsampled ($\frac{H}{8} \times \frac{W}{8}$) independent-view feature $\hat{F}t^{V{id}}$. The context across views $C_t^{V_{id}}$ is predicted from this feature via a lightweight temporal feature extractor $f_\phi$~\cite{sheng2022temporal}, where downsampling stages are removed to maintain spatial fidelity.
\begin{equation}
    C_{t}^{V_{id}} = f_\phi(\hat{F}_t^{V_{id}}).
\end{equation}
The resulting inter‑view context $C_t^{V_{id}}$ is fused with the bidirectional multi‑scale temporal contexts $C_t^{V_k,f_l}$ and $C_t^{V_k,b_l}$ (with $l\in\{0,1,2\}$) before being passed into the contextual encoder‑decoder. This integration enables the joint removal of temporal and inter‑view redundancies in a unified coding stage, while avoiding the computational burden of explicit disparity estimation.

\begin{figure}[t]
  \centering
   \includegraphics[width=\linewidth]{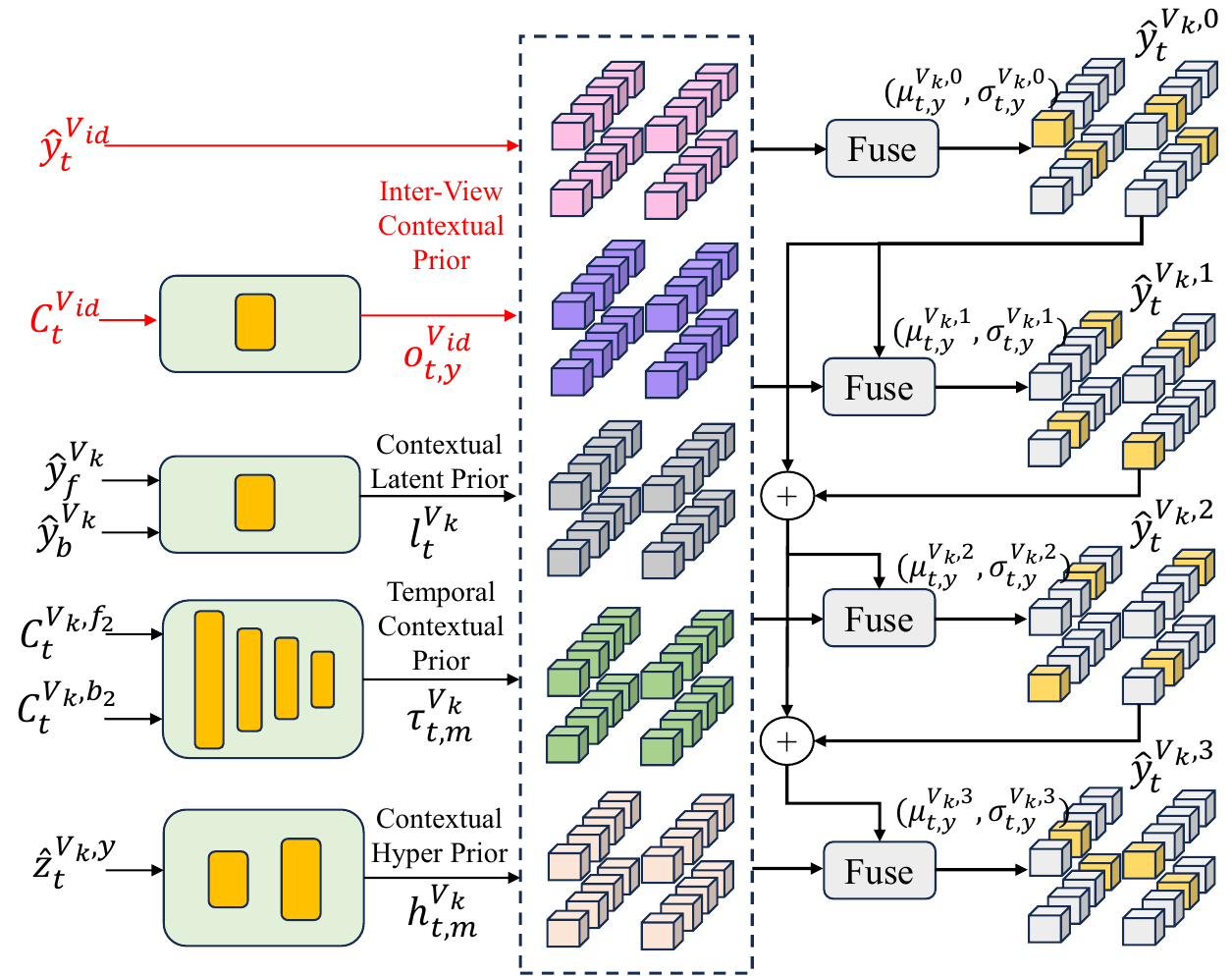}
      \caption{Contextual conditional entropy model with proposed inter-view context condtional priors. 
}
   \label{fig:ContextPrior}
\end{figure}
\subsubsection{Inter-View Contextual Conditional Entropy Model}\label{IVCCE}
To more effectively harness inter‑view content dependencies, we propose an inter-view contextual conditional entropy model by integrating inter-view contextual priors. As depicted in Fig.~\ref{fig:ContextPrior}, an inter-view contextual prior extractor $f_{\delta}$---composed of a convolutional layer---first processes the predicted inter-view context $C_t^{V_{id}}$ to produce an inter-view contextual conditional prior $o_{t,y}^{V_{id}}$:
\begin{equation}
    o_{t,y}^{V_{id}} = f_{\delta}(C_t^{V_{id}}).
\end{equation}
where $o_{t,y}^{V_{id}}$ shares the same spatial dimensions as the dependent-view content latent representation $\hat{y}_t^{V_k}$. Furthermore, we also propagate the content latent representation $\hat{y}_t^{V_{id}}$ of the independent view as another inter-view contextual conditional prior.

These inter-view contextual conditional priors are fused with the temporal contextual prior  $\tau_{t,y}^{V_k}$ predicted from the bi-directional temporal contexts ($\hat{C}_{t}^{V_k, f_2}$,$\hat{C}_{t}^{V_k, b_2}$), the contextual latent prior $l_t^{V_k}$ predicted from  bi-directional context latent representations ($\hat{y}_{f}^{V_k}$,$\hat{y}_{b}^{V_k}$), and the contextual hyper prior $h_{t,y}^{V_k}$ learned from the contextual hyper representation $\hat{z}_{t,y}^{V_k}$. The fused priors are fed into a spatial context model with quadtree partition~\cite{li2023neural} to estimate the distribution parameters ($\mu_{t,y}^{V_k,s}$, $\sigma_{t,y}^{V_k,s}$) of the $s^{th}$ dependent-view context latent representation segment $\hat{y}_t^{V_k,s}$:
\begin{equation}
p\left(\hat{y}_t^{V_k,s} \mid  o_{t,y}^{V_{id}}, \hat{y}_t^{V_{id}}, \tau_{t,y}^{V_k}, l_t^{V_k}, h_{t,y}^{V_k}, \hat{y}_t^{V_k, <s} \right) \sim \mathcal{L}\left(\mu_{t,y}^{V_k,s}, \sigma_{t,y}^{V_k,s}\right),
\end{equation}
where $\mathcal{L}\left(\mu, \sigma\right)$ denotes a Laplacian distribution and $\hat{y}_t^{V_k, <s}$ denotes previously compressed motion latent segments. 
\begin{figure}[t]
  \centering
   \includegraphics[width=\linewidth]{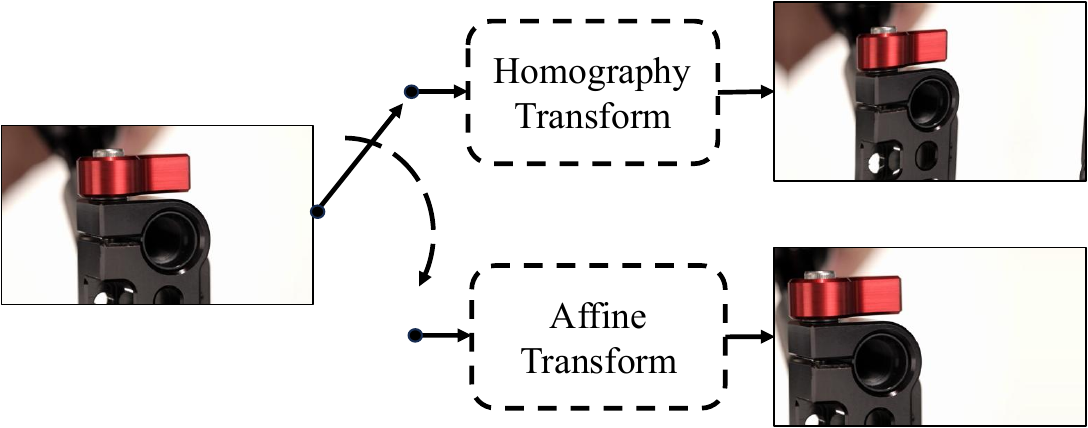}
      \caption{Proposed geometric transformation pipeline to synthesize multiview videos from single-view training videos.}
   \label{fig:view_synthesis}
\end{figure}
\begin{table}[t]
\caption{The multi-view test sequences used in our experiments.}
  \centering
\scalebox{1}{
\begin{threeparttable}
\begin{tabular}{c|c|c|c}
\toprule[1.5pt]
Test Videos    & 3 views    & 2 views & Video Resolution    \\ \hline
Balloons       & 1-3-5      & 1-3     & 1024$\times$768     \\ \hline
Newspaper1     & 2-4-6      & 2-4     & 1024$\times$768     \\ \hline
Kendo          & 1-3-5      & 1-3     & 1024$\times$768     \\ \hline
GT\_Fly        & 9-5-1      & 9-5     & 1920$\times$1088    \\ \hline
Shark          & 1-5-9      & 1-5     & 1920$\times$1088    \\ \hline
Undo\_Dancer   & 1-5-9      & 1-5     & 1920$\times$1088    \\ \hline
Poznan\_Hall2  & 7-6-5      & 7-6     & 1920$\times$1088    \\ \hline
Dance\_Dunhuang& 0-1-2      & 1-0      & 1920$\times$1080    \\ \hline
Dance\_Jasmine & 0-1-2      & 1-0      & 1920$\times$1080    \\ \hline
Kungfu\_Basic  & 0-1-2      & 1-0      & 1920$\times$1080    \\ \hline
Sport\_Taekwondo1& 0-1-2      & 1-0      & 1920$\times$1080    \\ \hline
Cadillac & 0-1-2      & 1-0     & 1920$\times$1080     \\ \hline
Carpark  & 0-1-2      & 1-0     & 1920$\times$1088     \\ \hline
Coffee\_Martini  & 7-8-9      & 8-7     & 1920$\times$1088    \\ \hline
Cut\_Beef  & 6-7-8      & 7-6     & 1920$\times$1088    \\ 
\bottomrule[1.5pt]
\end{tabular}
  \begin{tablenotes}
   \item \footnotesize \dag Coding order of 3-view videos: center-left-right~\cite{Rusanovskyy2013common}.
\item \footnotesize \ddag Coding order of 2-view videos: right-left~\cite{Rusanovskyy2013common}.
  \end{tablenotes}
\end{threeparttable}}
\label{table:testing_videos}
\end{table}
\begin{figure*}[t]
  \centering
  \begin{minipage}[c]{\linewidth}
  \centering
    \includegraphics[width=0.99\linewidth]{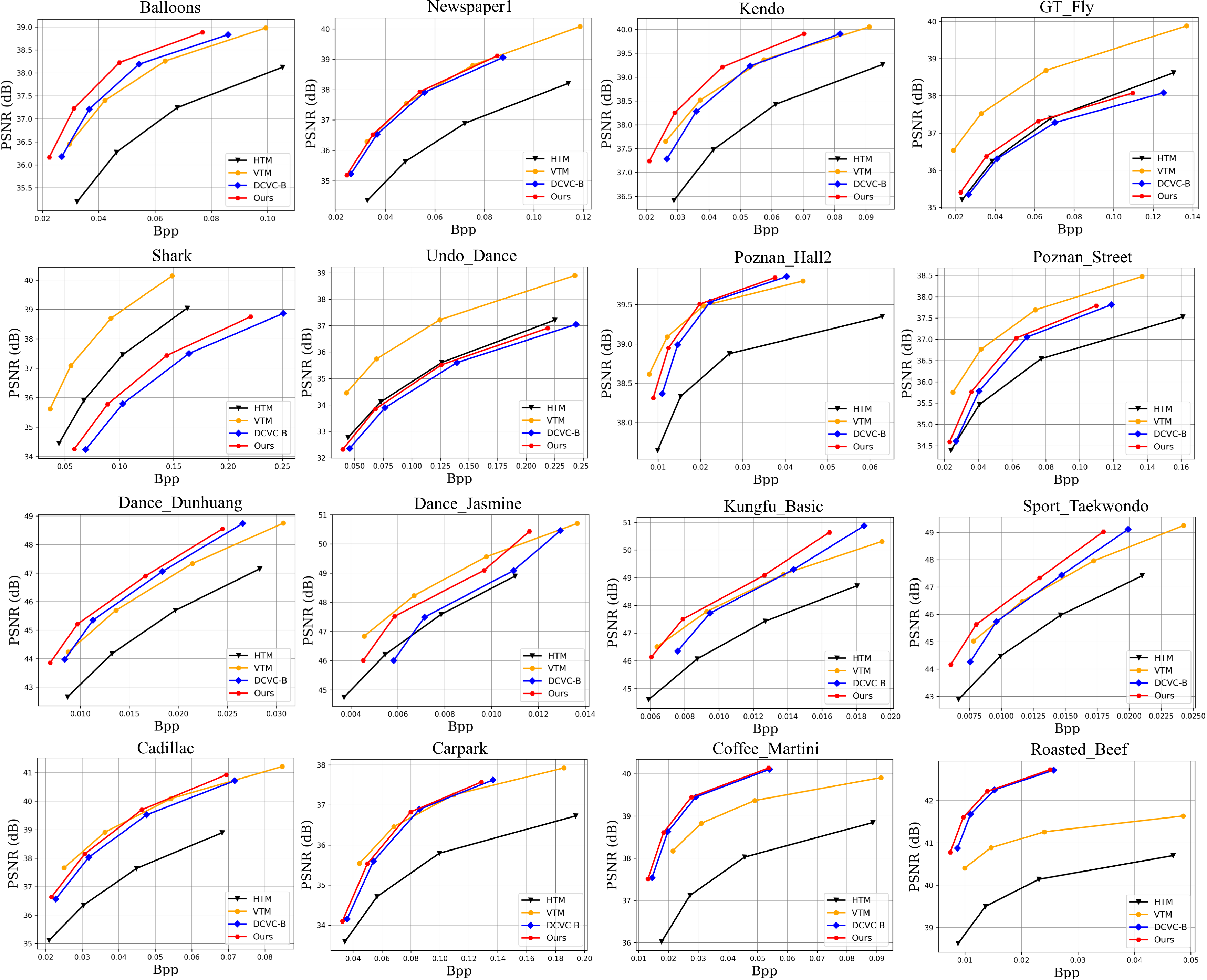}
  \end{minipage}%
    \caption{3-view video compression performance comparison of different codecs.}
  \label{fig:RDcurves_3view}
\end{figure*}


\begin{figure*}[t]
  \centering
  \begin{minipage}[c]{\linewidth}
  \centering
    \includegraphics[width=0.99\linewidth]{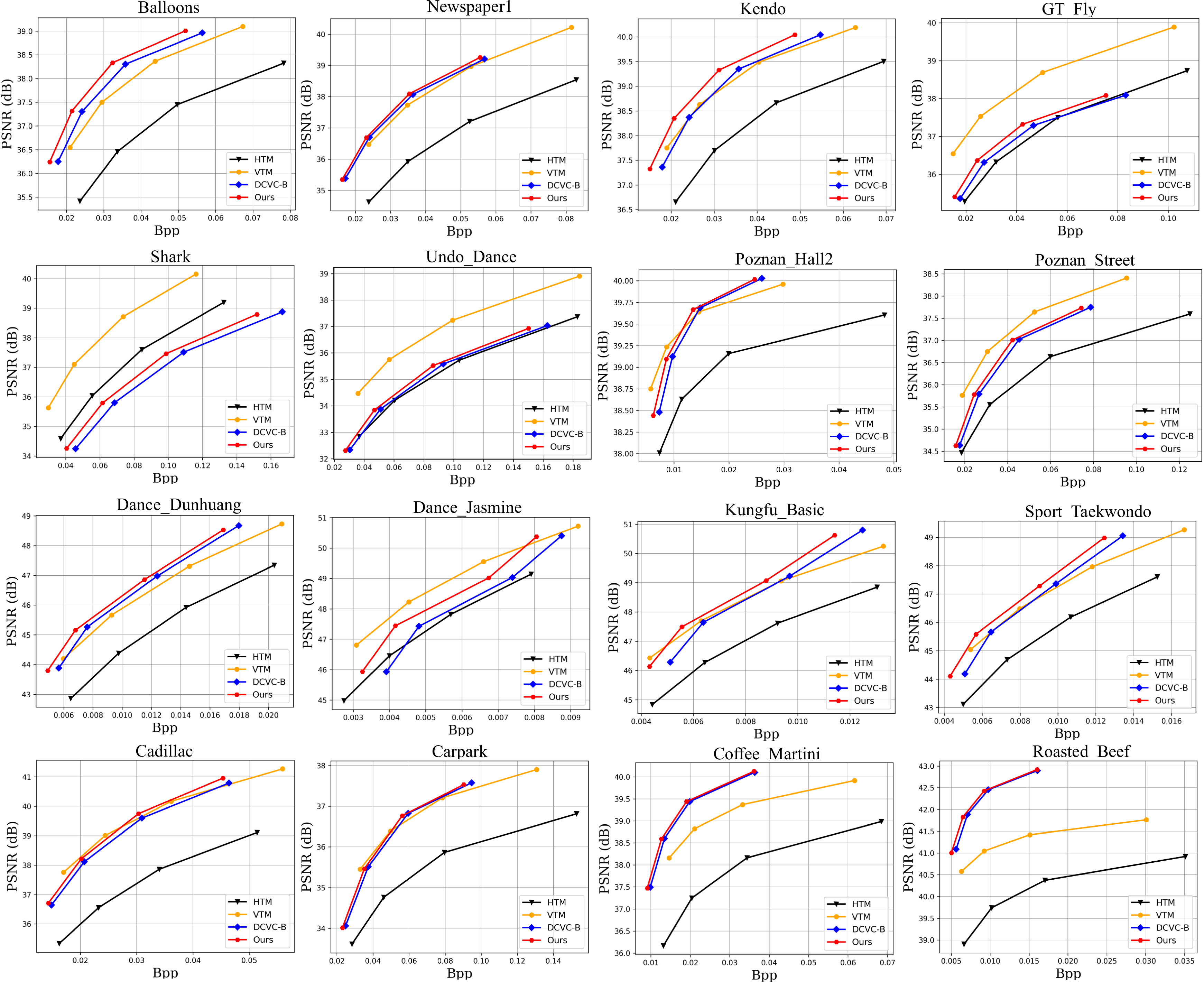}
  \end{minipage}%

    \caption{2-view video compression performance comparison of different codecs.}
  \label{fig:RDcurves_2view}
\end{figure*}
\begin{table}[t]
\caption{BD-rate (\%) comparison between our codec and HTM/VTM/DCVC-B on 3-view videos.}
  \centering
\scalebox{0.9}{
\begin{threeparttable}
\begin{tabular}{c|c|c|c}
\toprule[1.5pt]
                 & vs HTM & vs VTM & vs DCVC-B \\ \hline
Ballons          & --53.6 & --21.3 & --15.1    \\ \hline
Newspaper1       & --45.5 & --1.2  & --4.5     \\ \hline
Kendo            & --48.7 & --13.7 & --17.3    \\ \hline
GT\_Fly          & --8.1  & 114.5  & --15.3    \\ \hline
Shark            & 39.7   & 134.1  & --12.2    \\ \hline
Undo\_Dancer     & 4.2    & 98.7   & --8.3     \\ \hline
Poznan\_Hall2    & --56.8 & 4.6    & --11.6    \\ \hline
Poznan\_Street   & --30.4 & 34.5   & --9.9     \\ \hline
Dance\_Dunhuang  & --42.0 & --15.1 & --8.3     \\ \hline
Dance\_Jasmine   & --20.1 & 7.7    & --15.6    \\ \hline
Kungfu\_Basic    & --37.2 & --8.2  & --10.5    \\ \hline
Sport\_Taekwondo & --38.1 & --13.4 & --12.2    \\ \hline
Cadillac         & --40.2 & 0.6    & --7.2     \\ \hline
Carpark          & --42.1 & 0.2    & --5.4     \\ \hline
Coffee\_Martini  & --69.4 & --40.7 & --6.3     \\ \hline
Roasted\_Beef    & --85.2 & --63.8 & --8.1     \\ \hline
Average 		 & --35.8 & 13.6 & --10.5    
\\
\bottomrule[1.5pt]
\end{tabular}
\end{threeparttable}}
\label{table:3view}
\end{table}

\subsection{Training Strategy}\label{sec:synthesis}
Given the scarcity of real-world multiview video data, we employ a geometric transformation pipeline to synthesize multiview sequences from single-view training videos (see Fig.~\ref{fig:view_synthesis}). This pipeline employs a combination of affine and homography transformations to produce multiview data that maintains geometric consistency and realism.
To increase data diversity, the transformation parameters---including horizontal/vertical shifts and perspective intensity for homography, as well as horizontal displacement for affine transformation---are randomly varied within controlled ranges. This approach produces a rich set of training samples that effectively mimic the geometric consistency and viewpoint variations found in natural multiview content. \par

During training, the independent-view codec with frozen parameters first compresses the original single-view frames to obtain the independent-view representations. These representations are then used to condition the dependent-view codec, which is trained to compress the synthetically generated dependent-view frames. The dependent‑view codec is optimized with the following rate-distortion loss:
\begin{equation}
\begin{aligned}
    L_{t}^{V_k}&= w_t^{V_k} \cdot \lambda^{V_k} \cdot D_{t}^{V_k} + R_{t}^{V_k,m} + R_{t}^{V_k,y},
\end{aligned}
\label{loss}
\end{equation}
where $w_t^{V_k}$ denotes the hierarchical weighting factor~\cite{sheng2025bi} for different temporal layers, $\lambda^{V_k}$ is the Lagrangian multiplier that balances rate and distortion, $D_t^{V_k}$ measures the reconstruction error between the original and decoded dependent‑view frame, $R_t^{V_k,m}$ and $R_t^{V_k,y}$ denote the bit rates for motion and content coding, respectively.

\begin{figure*}[t]
  \centering
   \includegraphics[width=0.9\linewidth]{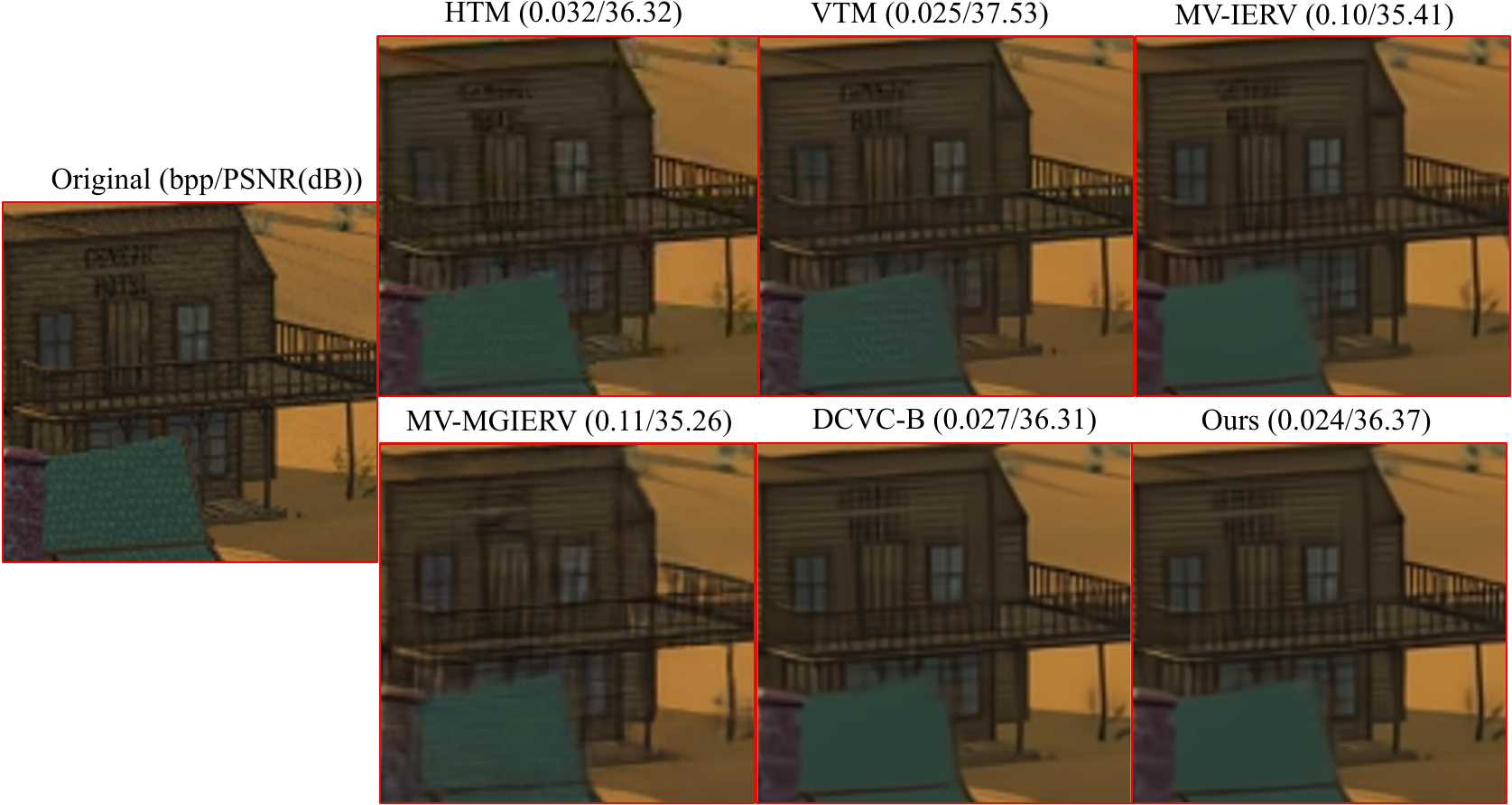}
      \caption{
Qualitative reconstruction quality comparison of the $9^{th}$ frame in View9 from GT$\_$Fly of different codecs.}
   \label{fig:subjective}
\end{figure*}

\begin{figure}[t]
  \centering
   \includegraphics[width=0.9\linewidth]{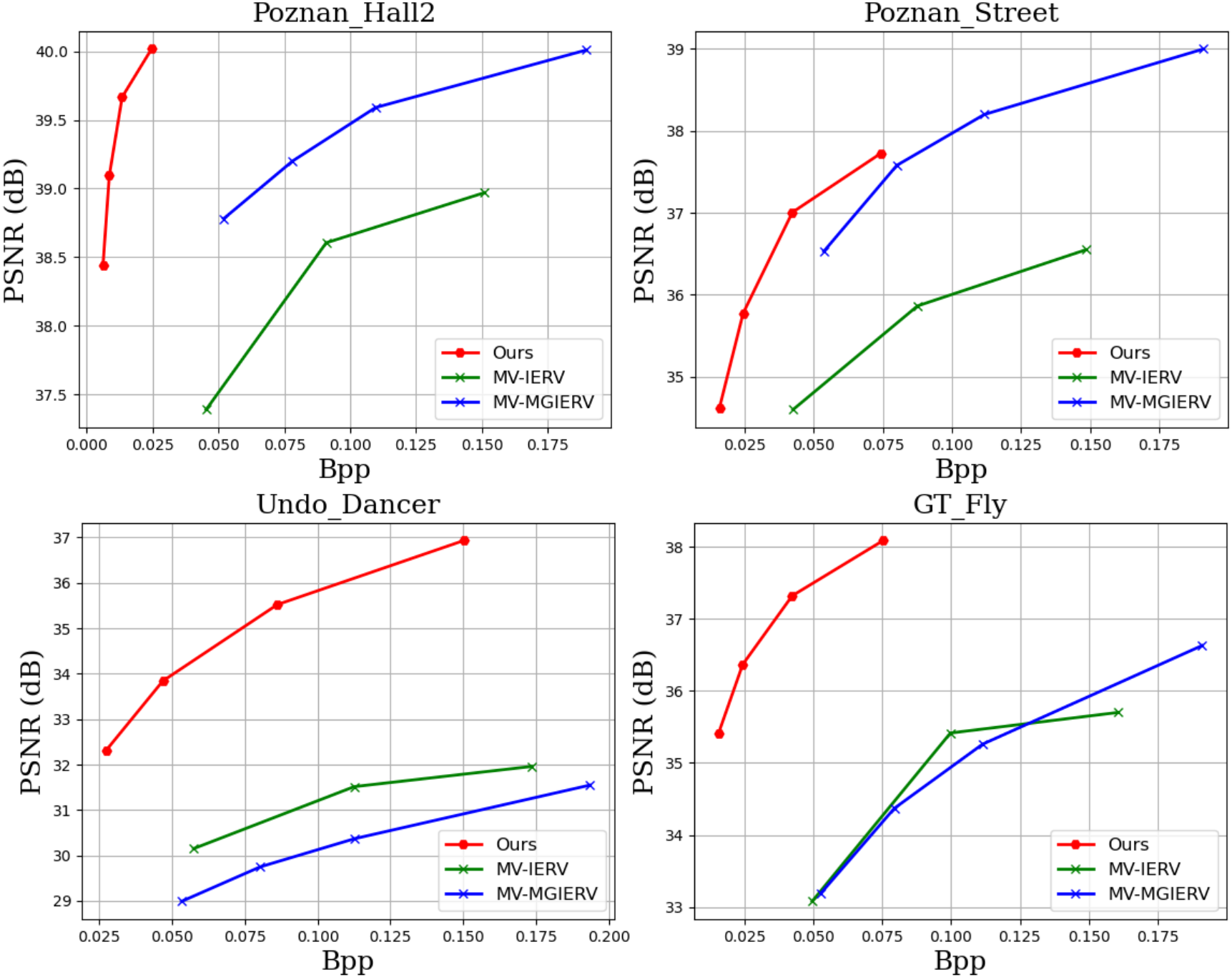}
      \caption{Compression performance comparison between our codec, MV-IERV~\cite{zhu2025implicit}, and MV-MGIERV~\cite{ling2025multi}.}
   \label{fig:INR}
\end{figure}

\section{Experiments}\label{sec:experiments}
\subsection{Experimental Setup}
\subsubsection{Datasets}
For training, multi‑view sequences are synthesized from single‑view videos using the geometric transformation pipeline described in Section~\ref{sec:synthesis}. The single‑view source videos are drawn from the Vimeo‑90k dataset~\cite{xue2019video}. For evaluation, as listed in Table~\ref{table:testing_videos}, we adhere to the standardized MV-HEVC common test conditions~\cite{Rusanovskyy2013common} and employ all of its official multi-view test sequences. In addition, to verify the generalization of our codec, we evaluate 8 videos from the PKU-DyMVHumans~\cite{zheng2024pku}, MPEG immersive video~\cite{boyce2021mpeg}, and N3DV~\cite{li2022neural} datasets. \par
\begin{table}[t]
\caption{BD-rate (\%) comparison between our codec and HTM/VTM/DCVC-B on 2-view videos.}
  \centering
\scalebox{0.9}{
\begin{threeparttable}
\begin{tabular}{c|c|c|c}
\toprule[1.5pt]
                 & vs HTM & vs VTM & vs DCVC-B \\ \hline
Ballons          & --54.5 & --22.3 & --11.6    \\ \hline
Newspaper1       & --48.2 & --8.8  & --3.7     \\ \hline
Kendo            & --47.0 & --14.0 & --13.2    \\ \hline
GT\_Fly          & --20.1 & 88.8   & --11.7    \\ \hline
Shark            & 20.8   & 98.1   & --9.2     \\ \hline
Undo\_Dancer     & --10.2 & 65.9   & --6.2     \\ \hline
Poznan\_Hall2    & --54.0 & 1.0    & --8.9     \\ \hline
Poznan\_Street   & --36.2 & 23.9   & --7.5     \\ \hline
Dance\_Dunhuang  & --40.4 & --12.9 & --6.2     \\ \hline
Dance\_Jasmine   & --15.1 & 13.7   & --11.7    \\ \hline
Kungfu\_Basic    & --35.9 & --6.8  & --8.3     \\ \hline
Sport\_Taekwondo & --36.4 & --10.7 & --9.1     \\ \hline
Cadillac         & --45.9 & --0.7  & --5.8     \\ \hline
Carpark          & --46.2 & --3.0  & --4.0     \\ \hline
Coffee\_Martini  & --69.6 & --39.4 & --5.1     \\ \hline
Roasted\_Beef    & --86.7 & --62.9 & --5.7     \\ \hline
Average 		 & --39.1 & 6.9 & --8.0    
\\
\bottomrule[1.5pt]
\end{tabular}
\end{threeparttable}}
\label{table:2view}
\end{table}

\subsubsection{Implementation Details}
To satisfy the random-access requirements, we adopt DCVC-B~\cite{sheng2025bi} with publicly available codes as our independent-view codec.
During training, we use Mean Squared Error (MSE) to measure the distortion. We set the Lagrangian multiplier $\lambda^{V_k}$ to ${85, 170, 380, 840}$, and set the hierarchical weight $w_t^{V_k}$ to ${1.4, 1.4, 0.7, 0.5, 0.2}$ for different temporal layers. We employ the AdamW optimizer~\cite{kingma2014adam} with a batch size of 8. To enable variable bit rate during testing, learnable quantization steps are embedded into both the motion and contextual auto-encoders of the dependent-view codec.  For comparison, the MV‑HEVC (multiview extension of H.265/HEVC) is implemented using its reference software HTM‑16.3 under the \texttt{baseCfg\_3view} and \texttt{baseCfg\_2view} configurations. In addition, we also compare with the H.266/VVC~\cite{bross2021overview} multiview profile, implemented with its reference software VTM‑21.2 under the \texttt{two\_layers} configuration. We set the total number of encoded frames to 97 and set the intra period to 32. \par

\subsubsection{Evaluation Metrics}
We use rate‑distortion (RD) analysis to evaluate the compression performance. We use the mean Peak Signal‑to‑Noise Ratio (PSNR) across all views to measure the reconstruction fidelity and use the total bits per pixel (bpp) averaged over all views to measure bit rate costs. We use the Bjontegaard Delta rate (BD-rate) to compare the compression performance of different codecs, where a negative value denotes better compression efficiency relative to the anchor method.

We use rate‑distortion (RD) analysis to evaluate the compression performance. We use the mean Peak Signal‑to‑Noise Ratio (PSNR) across all views to measure the reconstruction fidelity and use the total bits per pixel (bpp) averaged over all views to measure bit rate costs. Since end-to-end deep codecs with hyperprior structure usually involve downsampling and upsampling operations that require the spatial resolution to be a multiple of 64, we pad the input frames to meet this constraint. The PSNR is computed on the original resolution, and the total bits include those generated from the padded regions, while the bpp is calculated based on the original resolution. We use the Bjontegaard Delta rate (BD-rate) to compare the compression performance of different codecs, where a negative value denotes better compression efficiency relative to the anchor method.

\subsection{Experimental Results}
\subsubsection{Quantitative Comparison}
For 3‑view sequences, Fig.~\ref{fig:RDcurves_3view} and Table~\ref{table:3view} present the RD curves and BD‑rate results, respectively. DCVC‑MV yields considerable compression gains, attaining an average BD‑rate reduction of 35.8\% against the HTM‑16.3 anchor, with a standout reduction of 56.8\% on the Poznan$\_$Hall2 sequence. Relative to DCVC‑B, DCVC‑MV achieves 10.5\% BD‑rate reduction, confirming the efficacy of the proposed inter‑view prediction modules. For 2‑view sequences, the results in Fig.~\ref{fig:RDcurves_2view} and Table~\ref{table:2view} show a consistent trend. DCVC‑MV achieves a 39.1\% BD‑rate saving against HTM‑16.3.\par

Additionally, we compare DCVC‑MV with the multiview profile of H.266/VVC implemented in VTM‑21.2. On 3‑view and 2‑view sequences, DCVC‑MV obtains average BD‑rate increasing of 13.6\% and 6.9\%, respectively. For most natural content sequences, DCVC‑MV achieves superior or comparable performance to VTM. However, for computer‑graphics videos like GT\_Fly, Shark, and Undo\_Dancer, DCVC‑MV underperforms VTM. A detailed ablation study of this phenomenon is provided in Section~\ref{sec:CG}.\par

Furthermore, we compare DCVC‑MV with the state‑of‑the‑art implicit neural representation (INR) based methods MV‑IERV~\cite{zhu2025implicit} and MV-MGIERV~\cite{ling2025multi}. As shown in Fig.~\ref{fig:INR}, DCVC-MV achieves better compression performance against them.

\subsubsection{Qualitative Comparison}
Visual quality is assessed through enlarged regions of reconstructed frames, as shown in Fig.~\ref{fig:subjective}. Taking the $9^{th}$ frame of View 9 from GT\_Fly as an example, DCVC‑MV exhibits clearer textures, particularly along the edges of the green roof, and preserves finer details compared to other codecs, highlighting the perceptual benefits of the proposed inter‑view prediction mechanisms.
\begin{table}[t]
 \centering
 \caption{Model size, computational complexity, and encoding/decoding time comparison between DCVC-MV and DCVC-B.}
\scalebox{0.8}{
\begin{tabular}{c|c|c|c|c}
\toprule[1.5pt]
Schemes  & model size & MACs/pixel & Enc Time (GPU/CPU) & Dec Time (GPU/CPU) \\ \hline
DCVC-B  & 21.40M & 3004.52K & 0.50s (G)/67.04s (C) & 0.38s (G)/50.52s (C)\\ \hline
Ours    & 29.13M & 3457.38K & 0.57s (G)/105.84s (C)& 0.44s (G)/68.73s (C)\\
\bottomrule[1.5pt]
\end{tabular}}
\label{time}
\end{table}
\subsubsection{Model Size, Computational Complexity, and Runtime Comparison}
Table~\ref{time} compares the model sizes (the number of parameters), computational complexities, and runtime, obtained on a RTX3090 GPU and an Intel Xeon Gold 6330 CPU with dependent‑view frames of resolution $1024\times768$.
DCVC‑MV employs 29.13M parameters and consumes 3457.38K MACs/pixel, with an encoding time of 0.57s (GPU) / 105.84s (CPU) per frame and a decoding time of 0.44s (GPU) / 68.73s (CPU) per frame. In contrast, DCVC‑B utilizes 21.40M parameters, consumes 3004.52K MACs/pixel, and records encoding and decoding times of 0.50s (GPU) / 67.04s (CPU) and 0.38s (GPU) / 50.52s (CPU) per frame, respectively. These results indicate that DCVC‑MV delivers better compression performance while incurring only a moderate increase in computational complexity. In the future, we will further explore optimization strategies to reduce computational complexity while maintaining compression efficiency.

 \begin{table}[t]
\caption{Ablation studies of our proposed methods.}
\centering
\scalebox{1}{
\begin{tabular}{c|c|c|c|c|c}
\toprule[1.5pt]
Index         &IVMFP        & IVMCE        &IIVCP &IVCCE & BD-Rate (\%)\\ \hline
$I_0$&\XSolidBrush  &\XSolidBrush &\XSolidBrush &\XSolidBrush & 0.0  \\ \hline
$I_1$&\Checkmark    &\XSolidBrush &\XSolidBrush &\XSolidBrush &--3.8 \\ \hline
$I_2$&\Checkmark    &\Checkmark   &\XSolidBrush &\XSolidBrush & --5.3    \\\hline
$I_3$&\Checkmark    &\Checkmark   &\Checkmark   &\XSolidBrush &--6.9 \\\hline
$I_4$&\Checkmark    &\Checkmark   &\Checkmark   &\Checkmark   & --11.5    \\
\bottomrule[1.5pt]
\end{tabular}
}
\label{effectiveness}
\end{table}
\begin{figure}[t]
  \centering
   \includegraphics[width=0.9\linewidth]{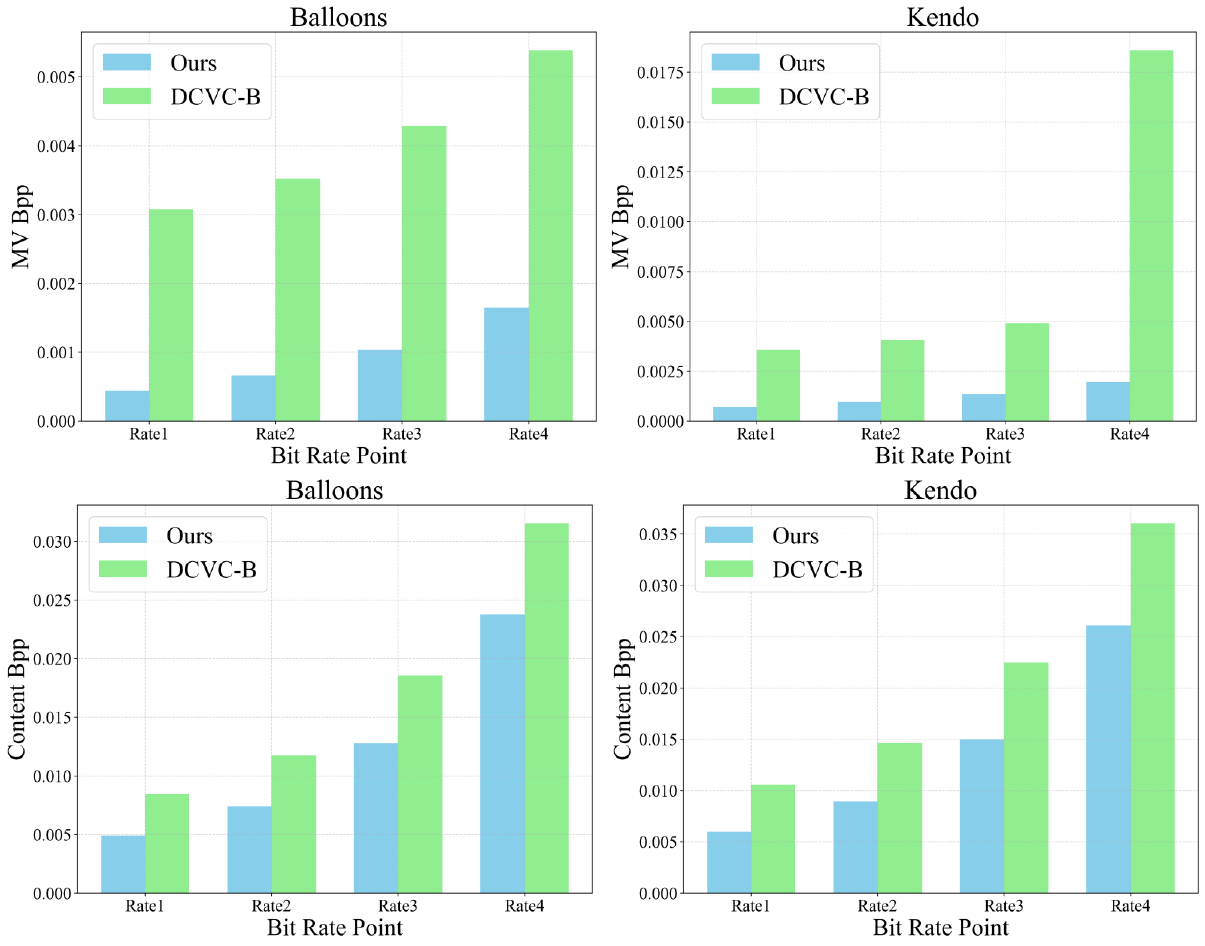}
      \caption{Bit rate comparison of motion vector and content bit rate between DCVC-MV and DCVC-B.}
   \label{fig:mv_bpp_comparison}
\end{figure}
\subsection{Ablation Study}
\subsubsection{Effectiveness of Proposed Techniques}
To assess the effectiveness of the proposed techniques, we carry out a systematic ablation analysis. Beginning with a baseline model $I_0$, the proposed components are integrated stepwise, yielding four successively improved models $I_1$–$I_4$ (see Table~\ref{effectiveness}). Comparing $I_0$, $I_1$, and $I_2$ confirms that the IVMFP method combined with the IVMCE model delivers a 5.3\% BD‑rate saving, demonstrating their ability to effectively reduce inter‑view motion redundancy. Extending $I_2$ with the IIVCP mthod and the IVCCE model in $I_3$ and $I_4$ further boosts compression efficiency, resulting in a total reduction of 11.5\% BD‑rate. The ablation results verify that IIVCP and IVCCE successfully leverage content dependencies across views to improve compression performance.

\subsubsection{Analysis of Inter-View Prediction}
To quantify the bit-rate savings enabled by the proposed inter-view modules, we analyze the rate distributions of DCVC‑MV and DCVC‑B for Balloons and Kendo with 3 views. DCVC‑MV consistently achieves lower bit rates across all operating points for both motion and content components.
For motion coding, DCVC-MV uses only 14.3\% of the bit rate of DCVC‑B at the lowest rate point on Balloons, increasing to 30.6\% at the highest rate. On Kendo, the relative motion ranges from 19.2\% to 32.8\%. For content coding, the bit‑rate proportion of DCVC‑MV relative to DCVC‑B spans 58.0\%-75.4\% on Balloons and 56.4\%–72.3\% on Kendo.
These results verify that the proposed inter‑view motion and context prediction methods significantly reduce the bit cost of both motion and content representations, which is a key contributor to the overall coding gain of DCVC‑MV.
\begin{figure}[t]
  \centering
   \includegraphics[width=\linewidth]{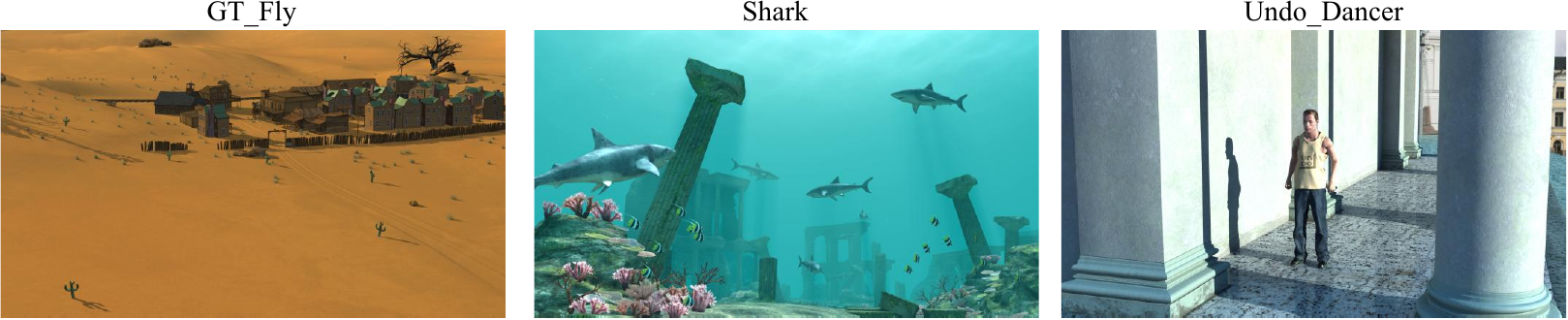}
      \caption{Illustration of computer-graphics test sequences.}
   \label{fig:bad_case}
\end{figure}
\begin{figure*}[t]
  \centering
   \includegraphics[width=0.9\linewidth]{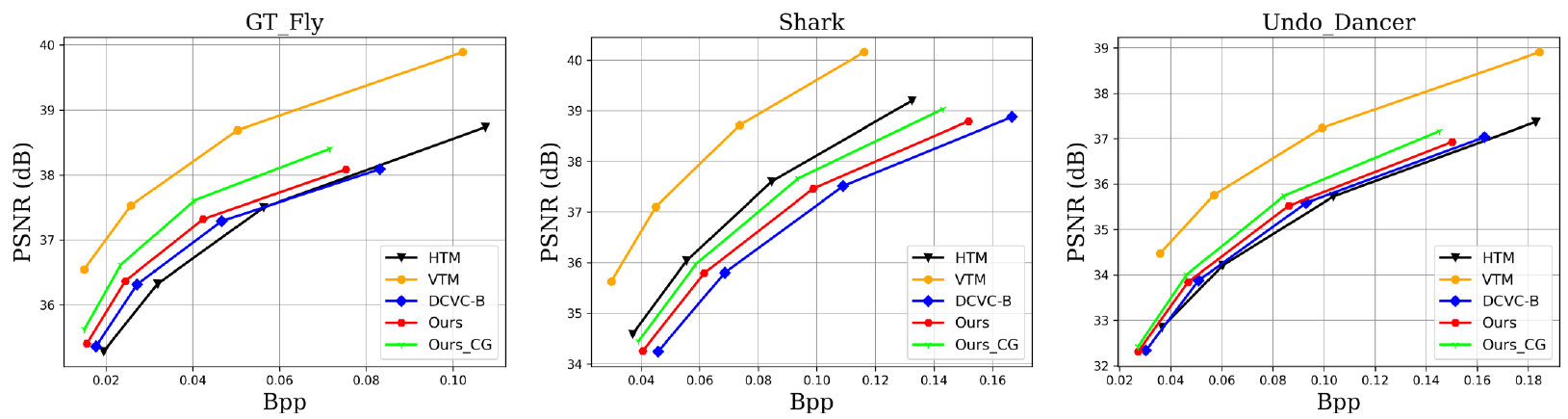}
      \caption{Compression performance variation of our codec on 2-view videos when incorporating computer-graphics videos into the training dataset.}
   \label{fig:CG_RD}
\end{figure*}

\subsection{Generalization to Computer-Graphics Video}~\label{sec:CG}
Our experimental evaluation also identifies a limitation regarding the compression of computer-graphics (CG) video. As shown in Table~\ref{table:3view}, DCVC‑MV notably underperforms HTM on animation sequences like Shark, with a BD-rate increase of 39.7\%. This performance shortfall, visualized in Fig.~\ref{fig:bad_case}, originates from a domain shift between the naturalistic video data used for model training and the distinct stylistic attributes present in these CG test sequences. Such behavior aligns with known challenges in learned video coding when facing out-of-domain videos, as also noted in prior work~\cite{sheng2022temporal}.\par

To mitigate this issue, we incorporate 1275 CG videos (each of 33 frames) into our training dataset and fine-tune our dependent-view codec. To ensure a fair comparison with DCVC‑B, we keep the independent-view codec unchanged. As shown in Fig.~\ref{fig:CG_RD}, the fine-tuned model (denoted as Ours\_CG) achieves substantial performance gains on CG sequences. Specifically, on GT\_Fly, Shark, and Undo\_Dancer, the BD-rate relative to HTM improves from -20.2\%, 20.9\%, and --17.6\% to --33.0\%, 9.0\%, and --17.6\%, respectively. These results demonstrate that incorporating CG data into training effectively alleviates the domain shift problem.\par

Nevertheless, a performance gap remains compared to traditional codecs such as VTM on CG content. This gap can be attributed not only to the limited diversity and scale of CG training data but also to inherent advantages of traditional codecs in handling CG sequences. For example, CG videos often contain large static backgrounds. Traditional codecs employ efficient skip modes and motion vector prediction to minimize overhead, whereas learned codecs still allocate bits to such regions.\par
We believe that further expanding the CG training dataset, jointly fine-tuning both dependent- and independent-view codecs, and designing specific technologies tailored to CG videos will further narrow this compression performance gap.

\section{Conclusion}\label{sec:conclusion}
This paper introduces DCVC-MV, a deep contextual multiview video compression framework that addresses the key practical demands of backward compatibility, random access, and effective inter-view prediction. The framework is built upon a structured hierarchical reference architecture that ensures the independent view is decodable as a single-view stream while supporting flexible viewpoint switching. To efficiently exploit cross-view dependencies, we introduce four key technical components: an inter-view motion feature propagation method and a corresponding inter-view motion conditional entropy model to capture motion correlation; and an implicit inter-view context prediction method paired with an inter-view contextual conditional entropy model to utilize content correlation. 
Experimental evaluations demonstrate that DCVC-MV delivers significant rate-distortion improvements relative to the conventional MV-HEVC standard and establishes a stronger baseline compared to independent single-view coding. As the first learned multiview codec that fully adheres to standardized structural constraints, DCVC-MV offers direct value to ongoing standardization efforts and facilitating future research in learned multiview compression.

\bibliographystyle{ieeetr}
\bibliography{ref}
\end{document}